\documentclass[sigconf]{acmart}

\AtBeginDocument{%
  \providecommand\BibTeX{{%
    \normalfont B\kern-0.5em{\scshape i\kern-0.25em b}\kern-0.8em\TeX}}}

\usepackage{subfigure}

\copyrightyear{2023}
\acmYear{2023}
\setcopyright{acmlicensed}
\acmConference[]{ }{}{}
\acmBooktitle{}
\renewcommand\footnotetextcopyrightpermission[1]{} 

\begin{document}
\pagestyle{plain} 

\title{Multimodal Temporal Fusion Transformers Are Good Product Demand Forecasters}

\author{Maarten Sukel}
\affiliation{%
  \institution{University of Amsterdam and Picnic Technologies}
  \city{Amsterdam}
  \country{Netherlands}}
  \email{m.m.sukel@uva.nl}

\author{Stevan Rudinac}
\affiliation{%
  \institution{University of Amsterdam}
  \city{Amsterdam}
  \country{Netherlands}}
  \email{s.rudinac@uva.nl}

\author{Marcel Worring}
\affiliation{%
  \institution{University of Amsterdam}
  \city{Amsterdam}
  \country{Netherlands}}
  \email{m.worring@uva.nl}
  
\renewcommand{\shortauthors}{Sukel, Rudinac and Worring}
\begin{abstract}
Multimodal demand forecasting aims at predicting product demand utilizing visual, textual, and contextual information. This paper proposes a method for multimodal product demand forecasting using convolutional, graph-based, and transformer-based architectures. Traditional approaches to demand forecasting rely on historical demand, product categories, and additional contextual information such as seasonality and events. However, these approaches have several shortcomings, such as the cold start problem making it difficult to predict product demand until sufficient historical data is available for a particular product, and their inability to properly deal with category dynamics. By incorporating multimodal information, such as product images and textual descriptions, our architecture aims to address the shortcomings of traditional approaches and outperform them. The experiments conducted on a large real-world dataset show that the proposed approach effectively predicts demand for a wide range of products. The multimodal pipeline presented in this work enhances the accuracy and reliability of the predictions, demonstrating the potential of leveraging multimodal information in product demand forecasting.
\end{abstract}

\ccsdesc[500]{Information systems~Multimedia and multimodal retrieval}
\ccsdesc[300]{Applied computing~Online shopping}
\ccsdesc[500]{Applied computing~Forecasting}

\keywords{demand forecasting, multimodal transformers, multimodal fusion}

\begin{teaserfigure}
  \includegraphics[width=\textwidth]{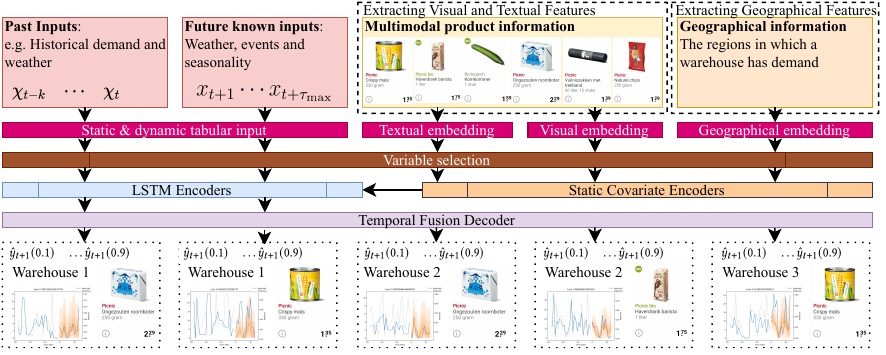}
  \caption{Tabular data sources such as weather, events, and seasonality are combined with embeddings of multimodal product information and geographical delivery information created using convolutional and graph-based methods in order to make multimodal product demand predictions using uses an encoder-decoder based transformer to handle static and dynamic information. Predictions are made for a range of quantiles per warehouse, delivery period, and product.}
  \label{fig:teaser}
\end{teaserfigure}

\maketitle

\section{Introduction}
Demand forecasting,  is a crucial task that has garnered extensive research in time-series analysis and regression analysis. Typical examples are predicting energy consumption \cite{bourdeau2019modeling,wei2019conventional}, sales trends, and various tasks in healthcare \cite{lim2018forecasting,zhang2018multi,alaa2019attentive}. The potential for forecasting in the economic sphere \cite{dong2020research}, such as traveling \cite{li2021multi} and particularly in the retail sector \cite{bose2017probabilistic}, is promising. The retail industry operates on thin margins, and goods are perishable, making an accurate forecast of product demand essential to optimize inventory management, minimize waste, and stay competitive. 

\begin{figure}
\centering
\includegraphics[width=\columnwidth]{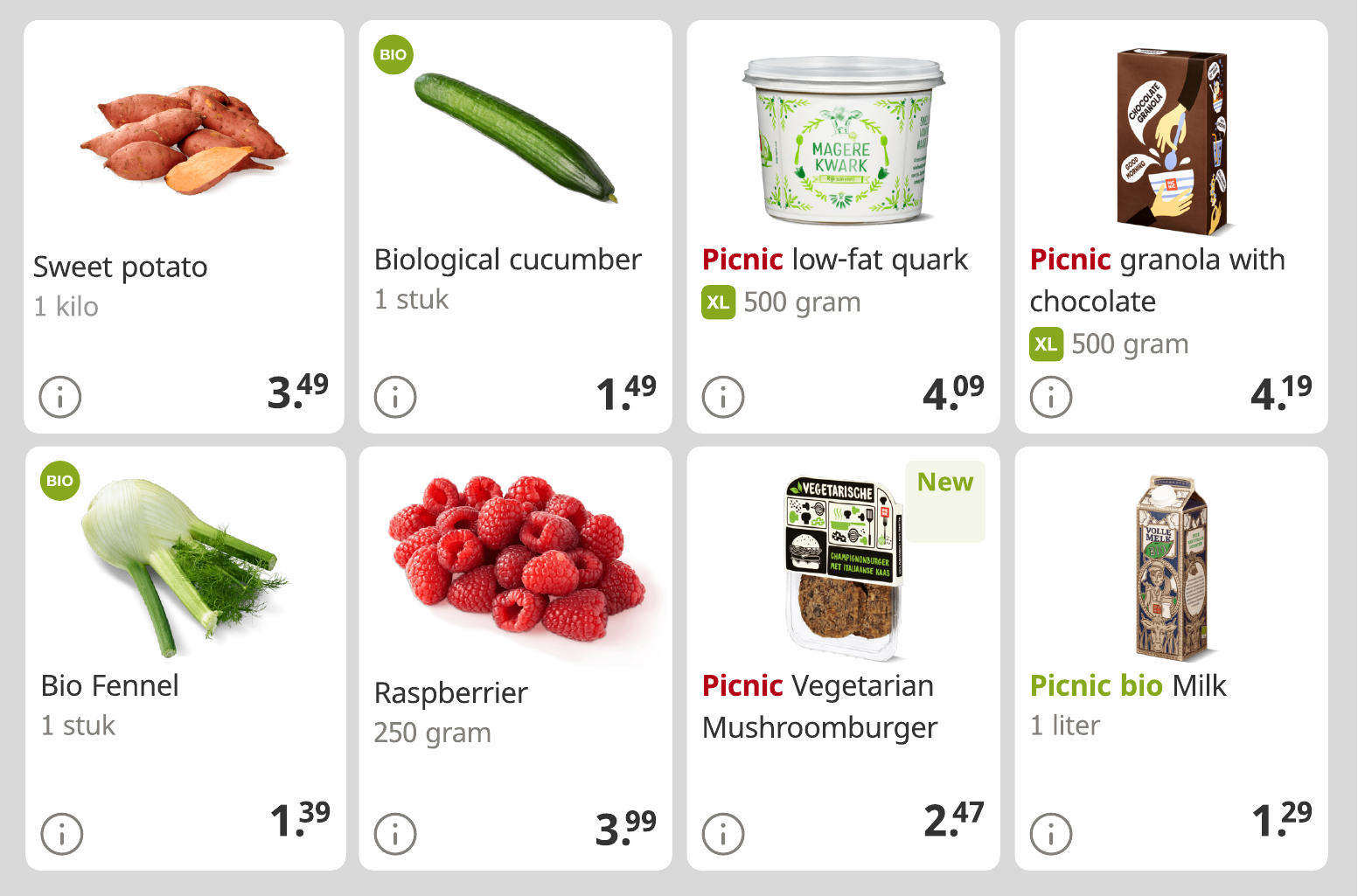}
\caption{Product images along with their textual description contain rich information that is used by consumers to decide what to purchase.}
\end{figure}

Product demand forecasting is a challenging task due to the large number of variables that come into play when predicting human behavior over the course of several weeks. Traditional demand forecasting methods often rely on tabular information like historical sales data, weather forecasts, events, and seasonalities. Even though these features are effective, they do not fully represent information about the type of product the demand is being predicted for or the context in which the product can be purchased.

In an online retail environment, customers base their purchase decisions on product images and textual descriptions \cite{wang2013influence}. Thus, visual and textual product information are decisive factors in potential product demand. There is abundant textual product information available for customers to decide what to buy, such as the name and description of the product, the nutrition label, and the list of ingredients. The visual aspect is important too for the customer \cite{crilly2004seeing}, since customers prefer appealing and properly depicted products \cite{10.1145/2964284.2967210}. Demand forecasting should consider both textual and visual product information. 

Geographic information is also an indicator of customer preference for specific types of products \cite{ramya2016factors}. Geographic context is particularly interesting for product demand forecasting because different regions may be associated with different customer behavior and consequently demand patterns. For example, customers in different regions may have distinct preferences for certain products due to differences in demographics, socio-economic factors, and type of housing. Weather, events, and seasonality have also been proven invaluable for understanding customer preference \cite{vergori2017patterns, fumi2013fourier, mirasgedis2006models}. 

To further improve product demand forecasting, tabular information should be combined with multimodal product information and geographical context about the delivery area. In this paper, we propose a novel multimodal approach to product demand forecasting, which achieves just that.

The task of product demand forecasting based on the different sources of information is challenging due to often having to predict multiple horizons into the future for a wide range of granularities. The granularities used in this work are \textit{warehouses}, \textit{products}, and different \textit{ordering moments} during the day. Product demand forecasting is demanding in a real-world setting due to the fast-changing assortment and high volumes of inference required to support supply chain operations. There are multiple reasons for the dynamic nature of categorical product information. For example, AB-testing and hyper-personalization are commonplace in such settings. Another common source of category dynamics is that the information about products is constantly optimized to improve the search, browsing, and filtering experience. The third reason is a frequent use of custom categories during special events, such as the holiday season. Due to the abrupt category changes and more gradual evolution, the data often contain noise. We conjecture that by relying more on product information and multimodal feature extraction methods, the levels and effects of noise in real word demand forecasting can be reduced. 

In this work, we create a multimodal method to extract features from products that are effective for product demand forecasting and fuse them with geographical and tabular information. Our approach is based on the use of a Temporal Fusion Transformer \cite{lim2021temporal}. Transformers have recently achieved state-of-the-art results in a wide range of natural language processing and computer vision tasks \cite{sanh2019DistilBERT,liu2022swin}, while also having state-of-the-art results in the domain of demand forecasting \cite{lim2021time,lim2021temporal} for tasks like predicting electricity consumption and traffic occupancy. In this paper, we make a significant step forward in multimodal demand forecasting by developing a Multimodal Temporal Fusion Transformer (MTFT) showing that by incorporating both product text and images as input modalities, a temporal fusion transformer is able to learn a more comprehensive representation of the product and its potential demand, while also capturing the preferences of customers in different geographic areas. For the textual product information, we propose a transformer-based component optimized for demand forecasting to extract the features from the product text, such as the name, description, and ingredients. For the extraction of visual features from product images we deploy a convolutional neural network-based component, optimized for generating input into the MTFT. The textual and visual approaches do not need large quantities of training data because their weights are pre-trained on large collections and fine-tuned afterwards. These multimodal features are then combined with the tabular features using Static Covariate Encoders and LSTM encoders in order to pass the information through the Temporal Fusion Decoder as described in Figure~\ref{fig:teaser}. The information extraction modules and the fusion layer form the core of our multimodal pipeline for demand forecasting. 

In order to study the role of different modalities, a large-scale real-world dataset of an online grocery store Picnic \cite{picnicengineering} is used for evaluation. Picnic operates in the Netherlands, Germany, and France. Picnic works by delivering groceries to people's houses the next day and operates out of several fulfillment centers that cover different geographical regions. Product demand forecasting is at the core of Picnic's operation and is used to accurately assess the quantity of all products customers will order. Due to the company being available online only, there is a large quantity of structured product information available in several different modalities. The data are used for an ablation study to investigate what multimodal information is beneficial for the tasks of product demand forecasting. In addition, different design choices for the architectures used for multimodal product demand forecasting are compared. The ablation study shows that a multimodal temporal fusion transformer can outperform a baseline using tabular features only.

The main contributions of this paper are as follows:

\begin{itemize}
    \item A state-of-the-art transformer-based approach for product demand forecasting that incorporates multimodal product description composed of textual and visual product information, and geographical demand information.
    \item The proposed approach does not require manually-generated product categories, making it more flexible and adaptable to real-world scenarios.
    \item An extensive evaluation on a large real-world dataset demonstrates the effectiveness and applicability of the proposed approach, paving the way for using multimodal approaches in product demand forecasting.
\end{itemize}

\section{Related work}
In this section we will first discuss traditional demand forecasting approaches, then the methods for extraction of unimodal and multimodal features, and finally, the work where multimodal approaches are deployed in related tasks.

\subsection{Demand forecasting}
Traditional approaches that extrapolate demand or use tree-boosting techniques have been effective in demand forecasting \cite{wolters2021joint,Chen:2016:xgboost}, but they can be lacking when the demand goes outside of the range found in the training set. However, advancements in deep learning have paved the way for more sophisticated techniques, including Temporal Fusion Transformers (TFTs) \cite{lim2021temporal}. They have been proven to be even more effective in accurately predicting demand in various fields, including retail, transportation, and finance. For longer-term forecasting, approaches like TiDe \cite{das2023long} have shown that an architecture based on multilayer perceptrons can achieve even better results than TFTs while having lower training cost and higher inference speed. The above-mentioned state-of-the-art demand forecasting approaches are not capable out-of-the-box of handling raw multimodal input. That is why in this work we develop approaches that can be used to feed multimodal features into a TFT model and evaluate them in a real-world application.

\subsection{Integration of multiple modalities}
In recent years, multimodal approaches have proven useful for a variety of tasks. For single modalities, like text in  natural language processing, there are transformer-based architectures such as DistilBERT  \cite{sanh2019DistilBERT} that achieve state-of-the-art results in a variety of tasks, such as question answering and text classification. In the visual domain, where tasks such as image classification and semantic segmentation were often performed using CNN-based approaches \cite{he2016deep}, transformers have also been achieving state-of-the-art results \cite{liu2022swin}. More recently, multimodal approaches improved a variety of tasks by being able to make effective representations that capture both visual and textual information. These multimodal approaches often unlock new possibilities. For example, combining visual and textual tasks allows for unified vision-language understanding and generation with approaches such as CLIP~\cite{radford2021learning} and BLIP~\cite{li2022blip}. In addition to unlocking new tasks, multimodal approaches can also improve performance on existing tasks. Examples are using geographical and also user click data to represent hotels \cite{sadeghian2019hotel2vec} or the forecasting of traffic flows \cite{he2022deep}. We believe that also in the case of product demand forecasting, incorporating multimodal features can result in improved performance and generalizability and multimodal transformers are a natural way to bring those information sources together.

\begin{figure*}
  \includegraphics[width=\textwidth]{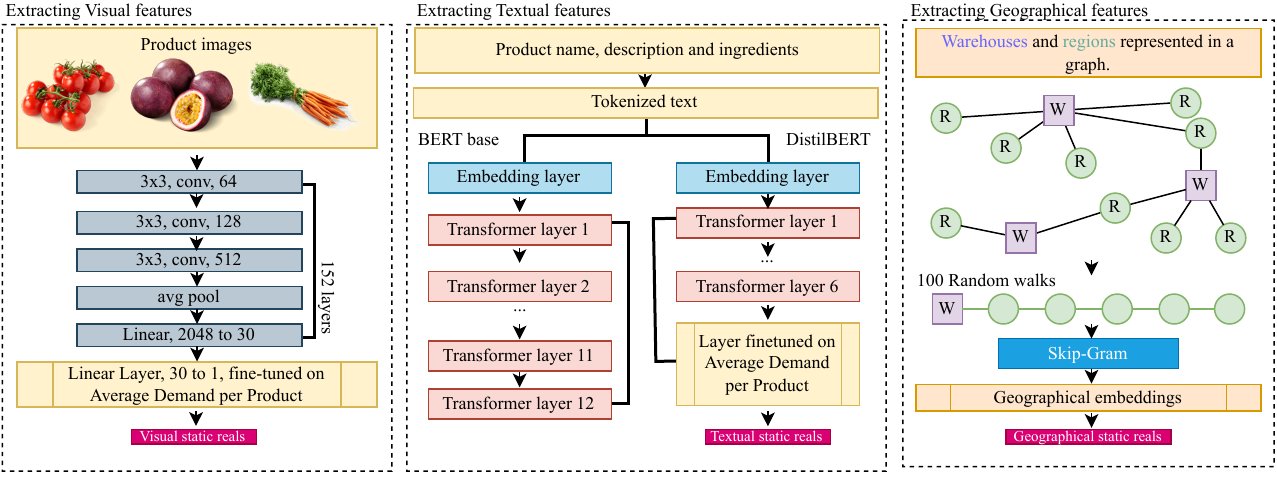}
  \caption{Feature extraction methods used in different parts of the proposed pipeline. The visual feature extraction is done using ResNet152~\cite{he2016deep} modified for demand forecasting. The textual features extraction utilizes DistilBERT~\cite{sanh2019DistilBERT} modified for demand forecasting. For the geographical embeddings, Node2Vec~\cite{grover2016node2vec} is used on a graph of warehouses and regions.}
  \label{fig:feature_extraction}
\end{figure*}

\subsection{Multimodal product representation learning}

Product embeddings can be used for various tasks such as search, recommendation, and personalization. Approaches for creating product embeddings like Data2vec \cite{pmlr-v162-baevski22a}, One Embedding To Do Them All \cite{singh2019one} or Prod2Vec \cite{vasile2016meta} are often used to improve recommender systems, while also showing potential in other tasks~\cite{singh2019one}.

Multimodal product analysis has been a popular research topic in multimedia. Food images and texts are used for extracting recipes and ingredient assessment tasks \cite{honbu2022unseen,yamakata2022recipe,wu2022ingredient,zhu2022cross,10.1145/3343031.3350574}. Product images are often used in information retrieval tasks and recommender systems such as the use of clothing images to help customers find relevant fashion \cite{gao2022fashion} or novel applications like screenless shopping \cite{gong2021inferring}. Due to the variety of products that occur in the fashion retail industry, retrieving the correct product for a customer is not a trivial task. Recent advancements in multimodal conversational search have allowed the user to more intuitively find the right products \cite{liao2018knowledge}. Multimodal deep neural networks are also used for attribute prediction and in e-commerce catalog enhancement \cite{sales2021multimodal}. In \cite{dheenadayalan2023multimodal} multimodal information from news articles is used for demand forecasting. This resulted in an increase in performance, demonstrating the potential of using multiple modalities to extract new and richer information for improving demand forecasting.  Based on these positive experiences with using multimodal product analysis in such applications, we conjecture it is likely that a multimodal approach for demand forecasting will yield good results. 

In conclusion, the use of multimodal product features for product demand forecasting is a relatively unexplored and exciting new field of research. With this paper, we aim to combine several novel multimodal feature extraction methods and evaluate their performance on large-scale real-world demand forecasting.

\begin{figure}
\centering
\includegraphics[width=0.46\textwidth]{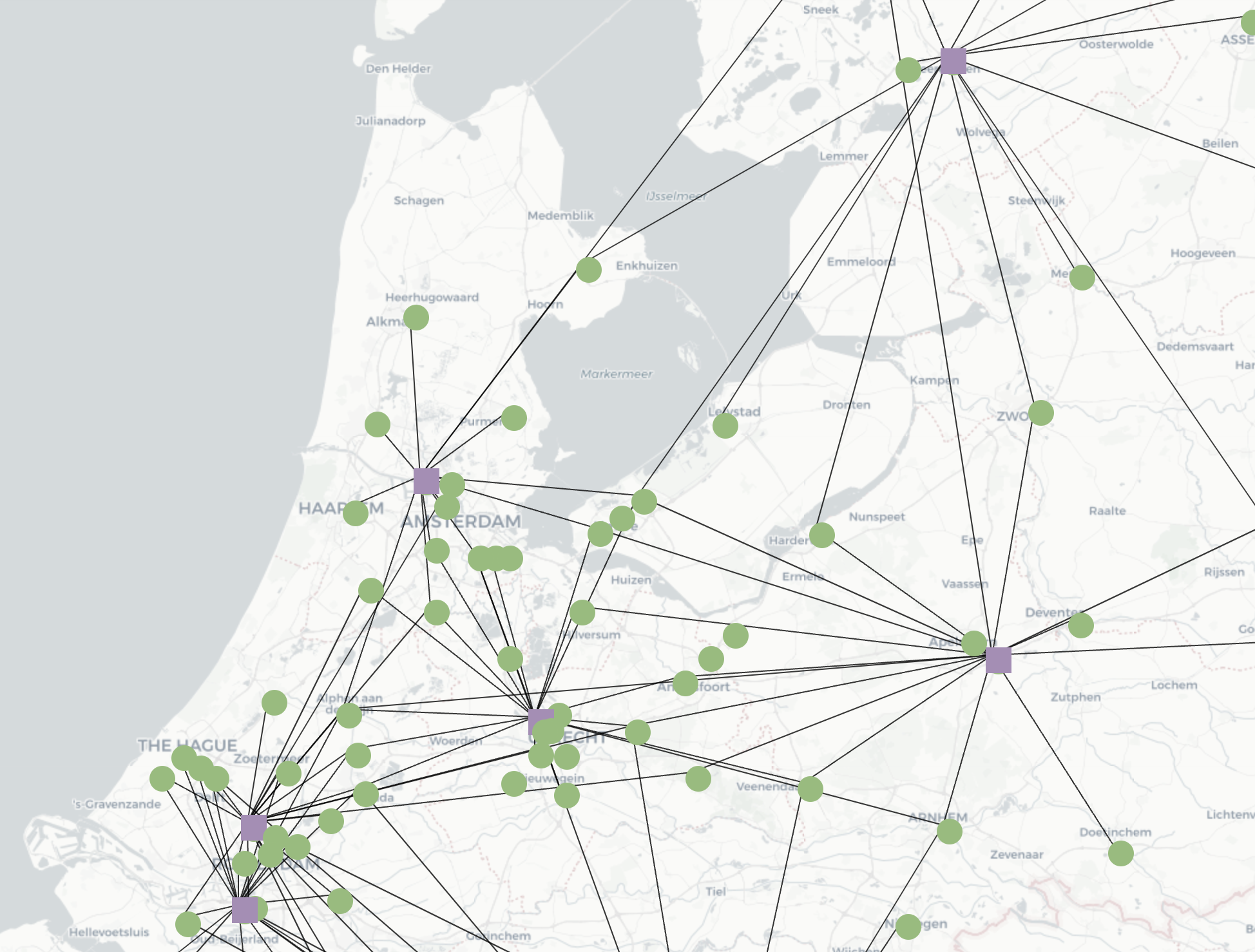}
\caption{Geographical overview of how the warehouses $\square$ are connected to specific regions $\circ$. For illustrative purposes, the number of regions has been reduced.
}
\label{figure:map}
\end{figure}

\section{Methodology}
In this section, we describe our proposed method for Multimodal Product Demand Forecasting. 

\begin{figure}
\centering
\includegraphics[width=\columnwidth]{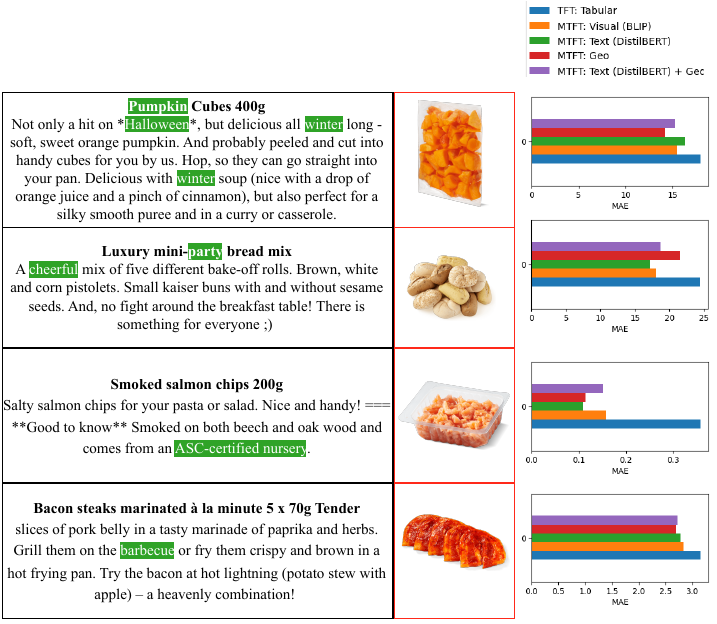}
\caption{Several examples of products where the textual and visual data can improve the product demand forecast, due to e.g., the availability of seasonal context.}
\label{fig:product_examples}
\end{figure}

There are several channels of information input into the pipeline. Visually, there are the images for every product $\mathcal{V} = \{v_1,v_2, ..., v_n\}$. Textually, for every product, there are names, descriptions, and ingredients $\mathcal{T} = \{(t_1^N, t_1^D, t_1^I),(t_2^N, t_2^D, t_2^I), ..., (t_n^N, t_n^D, t_n^I)\}$. Geographically, we have the regions $\mathcal{R} = \{r_1,r_2, ..., r_p\}$ and warehouses $\mathcal{W} = \{w_1,w_2, ..., w_q\}$. Finally, there is also traditional time $\tau$ dependent tabular information $\mathcal{I}_\tau= \{i_1,i_2, ..., i_s\}_\tau$. A detailed description of the variables is provided in Table~\ref{table:features} and in Figure~\ref{fig:product_examples} some further examples are given. The prediction output of the MTFT is at the level of the product, the warehouse, the delivery day, and part of the day. The predictions consist of quantiles, a way to estimate the range of possible outcomes for a given prediction, and can be defined as $ \hat{y}_{t+1}(0.1)\ldots\hat{y}_{t+1}(0.9)$. The traditional tabular information can be split up into past inputs $\chi_{t-k}\ldots\chi_t$ and known future inputs $\mathbf{x}_{t},\mathbf{x}_{t+1}, \ldots,\mathbf{x}_{t+\tau_{\max}}$. 

In the following sections, we describe how the input is constructed for the MTFT. Our approaches for feature extraction consist of four components. First, we present methods for creating visual and textual product embeddings by finetuning networks using demand forecasting as output and taking the learned representation. Afterward, we discuss the approach for the creation of geographical embeddings for the same task. Finally we consider the fusion of these different forecasts into one final prediction. 

\begin{table*}[htb]
\smaller
\caption{Overview of the traditional features used, the multimodal features and the actual.}
\label{table:features}
\begin{tabular}{lll}
\toprule
Feature   & Description & Example \\
\midrule
Time index & Integer with time index from prediction & -28 to 14 \\
Prediction date & The date of predicting & 2022-12-17\\
Prediction timestamp & The timestamp of predicting & 2022-12-17 12:01:00\\
Delivery date & The date of delivery & 2022-12-20\\
Delivery Weekday & The weekday of delivery & Monday\\
Delivery Month & The month of delivery & December\\
Warehouse & The warehouse of the forecast & UID\\
Part of day & Morning or evening delivery & Evening\\
Product & UID for product & UID\\
Products ordered at prediction time& Confirmed consumer units & 53\\
Confirmed number of deliveries at prediction time & The number of deliveries confirmed during prediction & 3550 \\
Rate of confirmed deliveries & The rate that deliveries are confirmed & 0.3\\
Minutes till slot cut off & The number of minutes till slot cut off& 42\\
Max temperature & Maximum temperature of the day & 3.1\\
Vacation period & If there is a vacation period in the delivery region & 0 \\
Christmas & Proximity till Christmas & 0\\
Text and Visual (BLIP) & Multimodal embedding created by using BLIP and PCA & Dimension of 10\\
Text (BLIP): Product Description, name, and ingredients & Textual embedding created by using BLIP and PCA & Dimension of 10\\
Visual (BLIP): Product image & Visual embedding created by using BLIP and PCA& Dimension of 10\\
Visual (ResNet): Product image & ResNet152 finetuned for product demand forecasting & Dimension of 10\\
Text (DistilBERT): Product Description, name, and ingredients  & DistilBERT finetuned for product demand forecasting & Dimension of 10\\
Geo (Node2Vec): Warehouses and regions & Geographical embedding created with node2vec & Dimension of 10\\
Number of product demand & The total consumer units, the target of the forecast& 203\\
\bottomrule
\end{tabular}
\end{table*}

\subsection{Extracting visual features} 
As explained in Introduction, the visual characteristics of a product are used by consumers to decide what purchases to make and are thus likely to influence demand. 

To investigate whether visual features can be used for the task of demand forecasting, we adjust ResNet512 \cite{he2016deep} for the tasks of product demand forecasting. In order to extract visual features, we have used a ResNet152 model pre-trained on ImageNet~\cite{deng2009imagenet}. An overview of the visual feature extraction is given in Figure~\ref{fig:feature_extraction}. Product images undergo preprocessing which includes adding padding to make them square, followed by resizing to 224x224. To diversify the dataset, a random horizontal flip and rotation are applied. The images are then normalized to match the pre-trained weights in ImageNet. The last layers of the ResNet are configured for linear output. The model is trained using Mean Squared Error Loss on the Average Product Demand, ensuring that the visual features extracted from the penultimate layer accurately represent the task of forecasting product demand. The dimensionality of the visual embedding is reduced to 10 by adding a linear layer to the ResNet architecture in order to allow it to go into the TFT as a static embedding. 

In addition to the feature creation method using ResNet, we also use a multimodal network, namely the BLIP \cite{li2022blip} architecture to extract visual features from product images. BLIP is a model trained on a range of textual and visual tasks, such as visual question answering, image-text retrieval, and image captioning, and is known for its capabilities to handle noisy input data in a wide range of domains including e-commerce and retail. Since the volume of data used by TFTs in a real-world setting is high, the dimensionality of the representations created with BLIP needs to be reduced. Therefore, we apply dimension reduction in order to make the dimensions feasible for the TFT with a dimensionality of ten. The method most effective for dimension reduction for these tasks found during experiments was Principal Component Analysis (PCA). 

\subsection{Extracting textual features}
Textual features of the products, such as the description, name, and ingredients are likely to have intrinsic details about the product. For example, if barbeque is mentioned in the description, a product is more likely to sell during the holidays and warm days, examples of which can are in Figure \ref{fig:product_examples}. But also the mentioning of certain certifications for salmon could be essential for a region where customers value sustainability over price. 

The DistilBERT architecture is deemed effective for the tasks of extracting textual information from multilingual sources, since it is smaller than comparable models and yet yields similar results. To adjust the textual embeddings for product demand forecasting, a pre-trained multilingual DistilBERT~\cite{sanh2019DistilBERT} model is utilized and a last layer is added with a smaller dimension than the original architecture in order to have a dimensionality that can be fed into the temporal fusion transformer as an embedding. A multilingual uncased tokenizer and multilingual uncased pre-trained model are used as a base. The model is then trained on the same target as the visual features, the average demand for a specific product. The textual features used include the product name, ingredients, and description. The dimensionality of the textual features is reduced to 10 by adding a linear layer to the final layer of the DistilBERT model. Figure \ref{fig:feature_extraction} provides an overview of the textual feature extraction pipeline.

Like for visual feature extraction, as an alternative, we also use the BLIP architecture to create a textual embedding. The same dimension reduction as for the textual approach is applied here.

We conjecture that the combination of visual and textual features should be able to capture the essence of products and improve the overall performance of the TFT for demand forecasting, and thus we also experiment with a combination of textual and visual input using BLIP, similar to the unimodal BLIP variants. For the creation of textual and visual features, the BLIP model \cite{li2022blip} has been used to create embeddings from textual product information and product images. 

\subsection{Extracting geographical features}
The area in which customers live has an influence on their buying behavior since they are likely to have different preferences \cite{ramya2016factors}. In order to allow the TFT model to capture this during training we set up a pipeline for the creation of geographical features. The products can have a certain geographical nature: for example, if Halloween is celebrated in a certain area, there may be more demand for pumpkin-related products as illustrated in Figure~\ref{fig:product_examples}.

For the geographical embeddings, we create a graph structure consisting of all the regions in the vicinity of a specific warehouse. The regions are areas where orders are delivered as can be seen in Figure \ref{figure:map}. The nodes of this graph consist of regions $\mathcal{R}$ and warehouses $\mathcal{W}$. The edges between these nodes represent if there are deliveries from the warehouse to the region. We employ a graph-based approach, utilizing Node2vec \cite{grover2016node2vec}, to transform this geographic region into effective demand forecasting representations, evaluating the feasibility of using geometric deep learning for this purpose. An overview of this approach can be seen in Figure \ref{fig:feature_extraction}. In Figure \ref{figure:map} a geographical schematic can be seen that displays the regions and warehouses location. The embeddings created are descriptive of the regions related to the delivery area of a specific warehouse, and pave the way for the TFT to learn about localized customer demand.

\begin{figure*}
    \centering
    \subfigure{\includegraphics[width=0.33\textwidth]{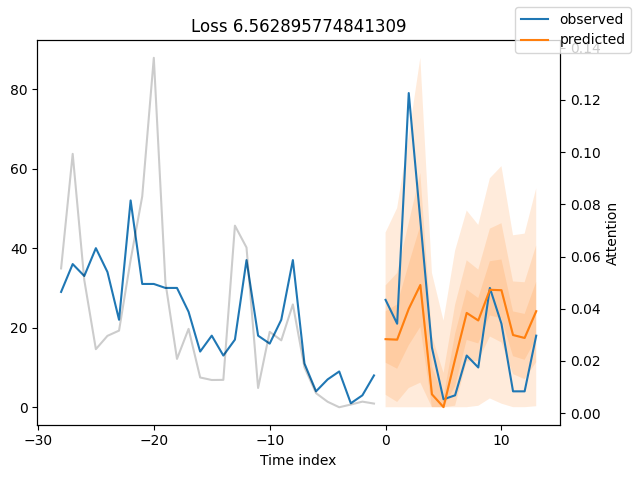}}
    \subfigure{\includegraphics[width=0.33\textwidth]{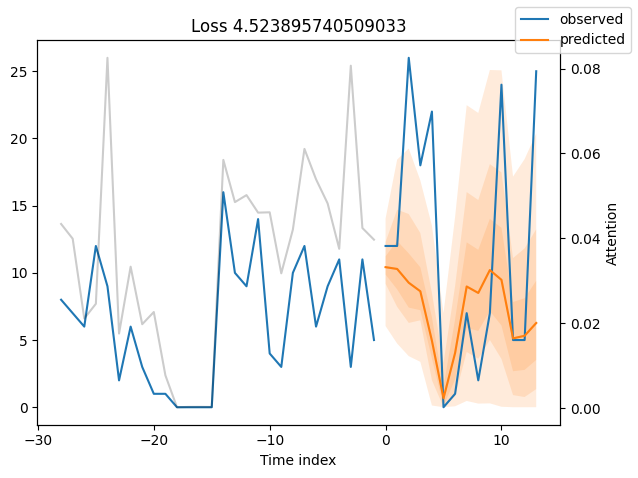}}
    \subfigure{\includegraphics[width=0.33\textwidth]{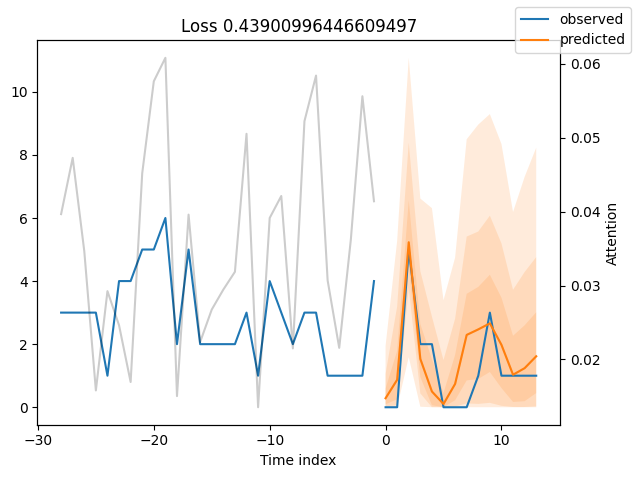}}
    \caption{For several products: Actuals in blue and the quantile prediction made for respectively product, day, warehouse, and part of the day in orange. The grey line shows the attention to that time index.
    }
\label{figure:five_images}
\end{figure*}

\subsection{Multimodal Temporal Fusion Transformer}
Two  configurations of a Multimodal Temporal Fusion Transformer have been trained. This will  give insight into what type of configuration works best. A MTFT is a Transformer-based model that leverages self-attention to capture the complex temporal dynamics of multiple time sequences and modalities, which makes it an effective tool for product demand forecasting allowing for a wide range of static and dynamic input. Dynamic features are, for example, reals like historical demand and weather, which change temporally or categorical information, such as the warehouse. Static input does not change over time, such as the product images and descriptions. Representations of the multimodal features are fused with the TFT model by incorporating them as embeddings based on the warehouse and the product creating the MTFT. These representations are combined with the past input and known future inputs. The output is at the warehouse level, date, parts of the day, and product. 

\section{Experimental setup}
In this section, we first discuss the data that has been used for the experiments and then the evaluation criteria used for the task of multimodal demand forecasting. 

\subsection{Data}
Dataset used for the experiment contains product texts and images for a wide range of products. The product texts include product descriptions, names, and ingredients, while the images depict segmented products. In addition to that we have several years of product demand to train and evaluate the model on. The dataset has the granularity of warehouse, product, delivery date, and delivery moment, which is a part of the day. 

In the context of demand forecasting for online grocery stores, one of the most important features is the information pertaining to confirmed orders. Given that orders are typically placed in advance for delivery within a specified time frame, it is possible to obtain knowledge about the number of specific products that have already been ordered for any given delivery moment. However, more traditional tabular data is also used, of which an overview can be found in Table \ref{table:features}.

The train (38.206.962 rows) and validation (5.643.705 rows) data used is from December 2021 to December 2022, and the evaluation is done on a test set (10.867.373 rows) of the two last weeks of 2022. It is to be noted that roughly 20\% of the dataset is used for evaluation, which is caused by sampling being more dense in this period making it a strong representation of how the pipeline would be applied in a real-world setting. By including the Christmas period in evaluation, we investigate how well the new features handle a period that is not trivial to predict demand for, while also taking the surrounding weeks into account, which can be considered as more regular. The dataset is not uniformly distributed, as some products have less demand than others. Also, a sparsity of products exists in the used dataset since not all products active in the test period are active in the (entire) train dataset. The idea is to capture the same dynamics that also occur in real-world environments.

\subsection{Evaluation criteria}
For the evaluation, we use the 0.5 output quantile and several well-known metrics, such as Mean Absolute Error (MAE, Eq.~\ref{eq:mae}), Mean Signed Deviation (MSD, Eq.~\ref{eq:msd}), and Root Mean Squared Error (RMSE, Eq.~\ref{eq:rmse}). This set of metrics captures a wide range of characteristics covering different aspects of demand forecasting. 
The metrics are also often used in the literature on the topic \cite{chicco2021coefficient}, where we add the MSD since over and under-forecasting are essential when predicting perishable products.  MSD can be used to indicate the deviation of the predictions in order to asses under or over-forecasting. The MAE is a good metric to evaluate the performance of a demand forecaster but is not as sensitive to outliers as the RMSE. Although sometimes used in related work\cite{dheenadayalan2023multimodal}, SMAPE is not included here due to a large number of zero values, which is not handled effectively by this metric. In the equations displayed below, $\hat{y}_i$ is the predicted value, $y_i$ is the true value and $n$ is the number of samples.

\begin{equation}
\label{eq:mae}
MAE = \frac{1}{n}\sum_{i=1}^{n}|\hat{y}_i-y_i|
\end{equation}

\begin{equation}
\label{eq:msd}
MSD = \frac{1}{n}\sum_{i=1}^{n} (\hat{y}_i - y_i)
\end{equation}

\begin{equation}
\label{eq:rmse}
RMSE = \sqrt{\frac{1}{n}\sum_{i=1}^{n}(\hat{y}_i-y_i)^2}
\end{equation}

\section{Experimental results}
In this section several experiments and their results are discussed, to answer the following research questions.

\begin{enumerate}
    \item Are Multimodal Temporal Fusion Transformers effective in multimodal product demand forecasting?
    \item What are the benefits of visual, textual, and geographical modalities, as well as the different pipelines for multimodal product demand forecasting?
\end{enumerate}

\subsection{Temporal fusion transformer architecture}
Due to the intricacy of the multimodal features, we evaluate what type of hyperparameters for the TFT would work best. We expect a smaller network not to be effective at capturing the complexity of these features. In order to evaluate that hypothesis, a large and a small network size have been configured as described in Table~\ref{table:hyper}. As can be seen in Table~\ref{table:small_results} when using a smaller network size there are improvements in some metrics, however, these improvements were not consistent for the range of metrics that we want to improve on.

In Table~\ref{table:large_results} it can be seen that with a larger network size, almost all the features in the ablation study improve performance. Especially the textual features extracted using the Textual and Geographical information embedded with Node2Vec improve performance on all metrics. It can thus be concluded that a large enough network is required for the effective usage of multimodal representations. With a large enough network size, Multimodal Temporal Fusion Transformers are effective at product demand forecasting.

\begin{table}[]
\small
\caption{Hyperparameters for the Temporal Fusion Transformers used for experimentation.}
\label{table:hyper}
\begin{tabular}{lllll}
\toprule
Model     & Hidden size & Continuous size & Dropout & LR \\
\midrule
Large TFT & 240         & 80                     & 0.2     & 0.001         \\
Small TFT & 24          & 16                     & 0.1     & 0.01         \\
\bottomrule
\end{tabular}
\end{table}

\subsection{Visual, textual and multimodal features for demand forecasting}
To understand the contribution of each modality on the overall performance of the model, we performed an ablation study, where we trained a Temporal Fusion Transformer with different combinations of unimodal and multimodal features. To assess the effectiveness of different approaches using visual, textual, and geographical information for demand forecasting, we performed experiments with a large and small model size of the Temporal Fusion Transformer.

\begin{table}[]
\small
\caption{The small TFT model trained and evaluated on 2662 products. Multimodal approaches do not yield consistently better results than the baseline when the network size is relatively small.}
\label{table:small_results}
\begin{tabular}{lrrrrr}
\toprule
      Run name &     RMSE &   MSD &  MAE \\
\midrule
TFT: Tabular &           8.40 & \textbf{-1.19} & \textbf{3.63} \\
MTFT: Text (BLIP)&       8.35 & -1.26 & 3.64 \\
MTFT: Visual (BLIP) &  8.56 & -1.38 & 3.64 \\
MTFT: Multimodal (BLIP)&   \textbf{8.01} & -1.20 & 3.66 \\
\bottomrule
\end{tabular}
\end{table}

\begin{table}[]
\small
\caption{The large TFT model trained and evaluated on 2662 products. Most multimodal approaches overperform the baseline model on multiple metrics.}
\begin{tabular}{lrrrrr}
\toprule
Run name &  RMSE &   MSD &  MAE \\
\midrule
MTFT: Text (DistilBert) + Geo &\textbf{7.74} & -1.06 & \textbf{3.35} \\

MTFT: Geo   &  7.85 & -0.92 & 3.36 \\
MTFT: Text (DistilBERT) &  7.76 & \textbf{-0.89} & 3.42 \\
MTFT: Visual (BLIP)&    7.98 & -1.35 & 3.43 \\
MTFT: Visual (ResNet) &   8.29 & -1.22 & 3.47 \\
MTFT: Visual (BLIP) + Text (BLIP)&  7.97 & -0.96 & 3.49 \\
MTFT: Text (BLIP)&  8.32 & -1.18 & 3.55 \\
TFT: Tabular &   8.29 & -1.30 & 3.56 \\
\bottomrule
\end{tabular}
\label{table:large_results}
\end{table}

\begin{figure}
\centering
\includegraphics[width=\columnwidth]{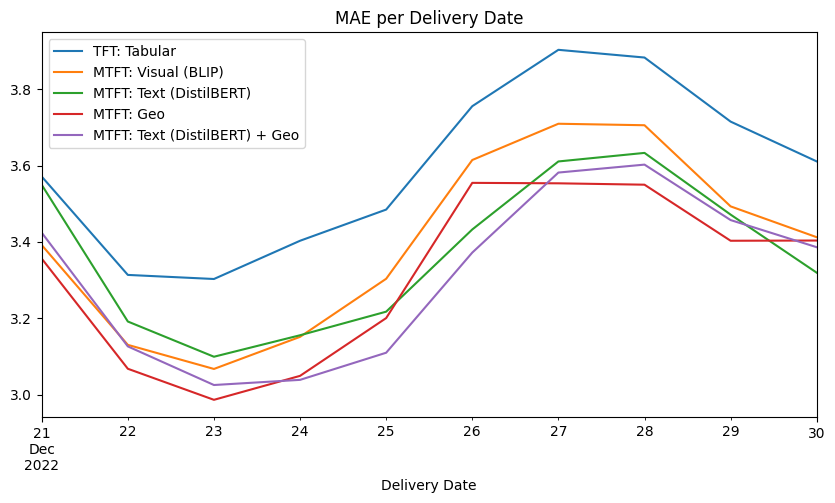}
\caption{MAE per delivery date showing the TFT baseline and the different MTFT variants.}
\label{fig:day}

\end{figure}
\begin{figure}
\centering
\includegraphics[width=\columnwidth]{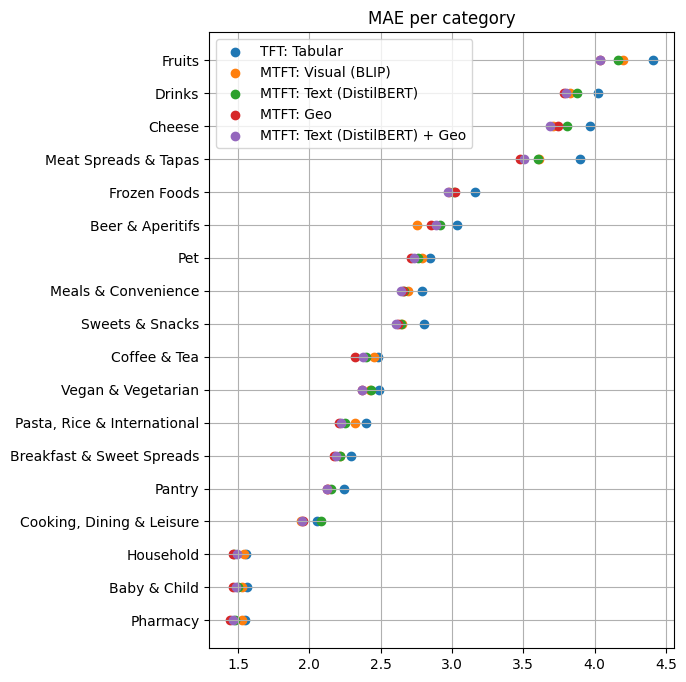}
\caption{MAE for a selection of product categories (more to the left is better), shows an improvement on almost all categories of products.}
\label{fig:cat}
\end{figure}

\begin{figure}
\centering
\includegraphics[width=\columnwidth]{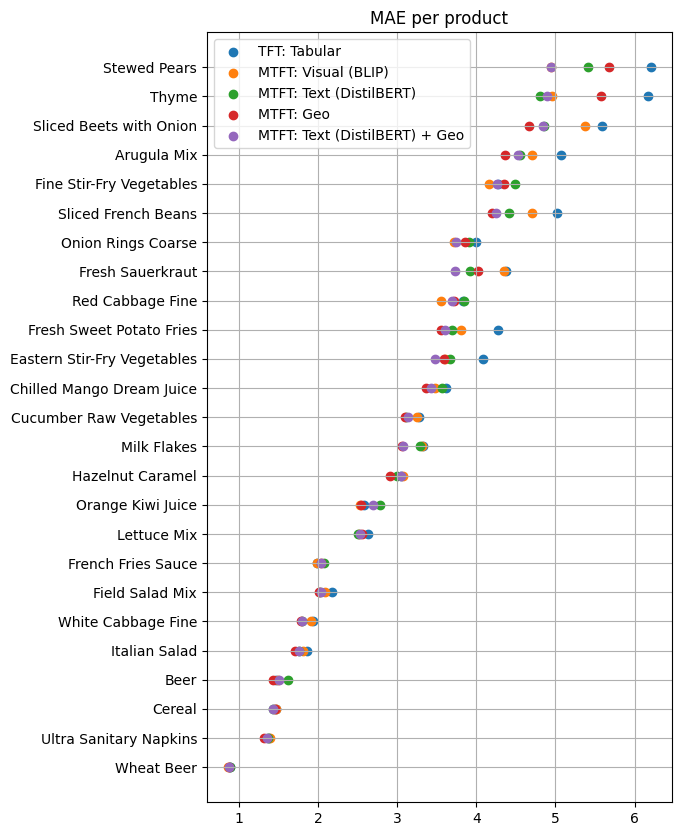}
\caption{MAE for a random selection of products.}
\label{fig:per_product}
\end{figure}

Results show that using multimodal information about products can improve the overall performance of demand forecasting.  The larger network size results are displayed in Table~\ref{table:large_results} and show that in a network with a large enough hidden size combined with multimodal features, a performance gain is obtained for all metrics:  RMSE, MSD, and MAE. The MSD metric shows that all approaches have the tendency to under-forecast. As can be seen in Figure~\ref{figure:five_images} utilizing higher output quantiles can be a solution for when there is a preference for under- or over-forecasting.

In Figure \ref{fig:day} it can be seen that the MTFT outperforms the baseline on all the delivery dates in the test set. The weekly pattern in performance that can be observed is due to promotions not being taken into account in this dataset and weekly demand not being equally distributed.

The MTFT outperforms the baseline in almost all categories. In figures \ref{fig:cat} and \ref{fig:per_product}, MAE score is displayed for a variety of categories and products. Categories such as `Fruits', `Drinks', `Meat Spreads \& Tapas' show the bigger improvements in MAE. A possible explanation for this is that these types of products have strong characteristics that are preferred by customers in a certain region.

\section{Conclusion}
In this paper, we proposed a novel multimodal approach to product demand forecasting that utilizes product texts, product images, and geographical embeddings as input modalities. The use of transformer-based neural networks allowed for the effective integration of textual and geographical information, leading to improved performance compared to traditional approaches. Our experiments on a novel real-world dataset demonstrate the effectiveness of the proposed approaches in predicting demand for a wide range of products. Additionally, our approach can work with the often noisy categorical product information. It handles the cold start problem of a new product by using visual and textual modalities which could allow for quicker adoption of newly introduced products. Additionally, our approach outperforms state-of-the-art baselines. Due to the scale of an online retailer, even small forecasting improvements result in an enormous reduction in the waste of products, while keeping sufficient products in stock. The ablation study also created new insights into the multimodal approaches that help to improve the performance of the model by utilizing the complementary information provided by the different modalities. Utilizing BLIP for the creation of multimodal features shows an improvement compared to the baseline when architecture is correctly selected, yet adopting transformer and convolutional-based models and fine-tuning them on the task of product demand forecasting yielded better results. The most effective features are textual and geographical, but also visual information can be used to improve demand forecasting. In conclusion, the use of multimodal product information and geographical embeddings is effective for the tasks of product demand forecasting. Finaly, since the retail sector is such a high-volume market with perishable goods, this work has a high potential for a positive impact on the environment and economic benefits to retailers, whilst paving the way for research into multimodal product demand forecasting.

\bibliographystyle{splncs04}
\bibliography{lib}
\end{document}